\pdfoutput=1

\documentclass[11pt]{article}

\usepackage[final]{acl}

\usepackage{times}
\usepackage{latexsym}

\usepackage[T1]{fontenc}

\usepackage[utf8]{inputenc}

\usepackage{microtype}

\usepackage{inconsolata}

\usepackage{graphicx}

\usepackage{algorithm}
\usepackage{algorithmicx}
\usepackage{makecell}
\usepackage{tabularx}
\usepackage{ragged2e}
\usepackage{multirow}
\usepackage{subfigure}
\usepackage{booktabs}
\usepackage{arydshln}
\usepackage{amssymb}  
\usepackage{pifont}   
\usepackage{amsmath}
\usepackage{algpseudocode}
\usepackage{etoolbox}
\usepackage[most]{tcolorbox}
\newtcolorbox{mybox}[1][]{
breakable,
  arc=1mm,
  boxrule=1pt,
  colback=yellow!14,
  colframe=black!80,
  fonttitle=\bfseries,
  title=#1,
  left=1mm,
  right=1mm,
  top=1mm,
  bottom=1mm
}
\usepackage{xcolor}

\title{The GaoYao Benchmark: A Comprehensive Framework for Evaluating Multilingual and Multicultural Abilities of Large Language Models}

\author{Yilun Liu$^1$\thanks{Equal contribution.}, Chunguang Zhao$^1$$^\ast$, Mengyao Piao$^1$, Lingqi Miao$^1$, Shimin Tao$^1$,\\ 
\textbf{Minggui He$^1$, Chenxin Liu$^1$, Li Zhang$^1$, Hongxia Ma$^1$, Jiaxin Guo$^1$, Chen Liu$^1$,}\\
\textbf{Liqun Deng$^1$, Jiansheng Wei$^1$, Xiaojun Meng$^1$, Fanyi Du$^1$,} \\
\textbf{Daimeng Wei$^1$, Yanghua Xiao$^2$} \\
$^1$ Huawei, China \\ 
$^2$ Fudan University, China \\ 
\texttt{liuyilun3@huawei.com, zhaochunguang6@huawei.com}
} 

\begin{document}

\maketitle

\begin{abstract}
Evaluating the multilingual and multicultural capabilities of Large Language Models (LLMs) is essential for their global utility. However, current benchmarks face three critical limitations: (1) fragmented evaluation dimensions that often neglect deep cultural nuances; (2) insufficient language coverage in subjective tasks relying on low-quality machine translation; and (3) shallow analysis that lacks diagnostic depth beyond simple rankings. To address these, we introduce \textbf{GaoYao}\footnote{GaoYao is derived from Chinese mythology, where he served as the first judicial officer, symbolizing fairness and comprehensiveness.}, a comprehensive benchmark with 182.3k samples, 26 languages and 51 nations/areas. First, GaoYao proposes a unified framework categorizing evaluation tasks into three cultural layers (General Multilingual, Cross-cultural, Monocultural) and nine cognitive sub-layers. Second, we achieve native-quality expansion by leveraging experts to rigorously localize subjective benchmarks into 19 languages and synthesizing cross-cultural test sets for 34 cultures, surpassing prior coverage by up to 111\%. Third, we conduct an in-depth diagnostic analysis on 20+ flagship and compact LLMs. Our findings reveal significant geographical performance disparities and distinct gaps between tasks, offering a reliable map for future work. We release the benchmark\footnote{https://github.com/lunyiliu/GaoYao}.
\end{abstract}

\section{Introduction}
\label{sec:introduction}

As Large Language Models (LLMs) increasingly serve a global user base, the ability to process diverse languages and navigate complex cultural contexts has become a critical measure of their inclusivity. However, the current landscape of multilingual evaluation is fraught with challenges that hinder a holistic understanding of model performance:

\textbf{(1) Lack of Systematicity and Cultural Neglect.} Many prominent benchmarks focus narrowly on single specific facets of language ability, such as factual knowledge~\citep{romanouinclude} or reading comprehension~\citep{bandarkar-etal-2024-belebele}. Consequently, they often overlook the deeper capabilities a model should possess (\emph{e.g.}, cultural sensitivity), treating multilingualism merely as isolated evaluation points rather than interconnected dimensions rooted from cultural and cognitive sources.

\textbf{(2) Limited Language Coverage and Quality in Subjective Tasks.} Subjective tasks (\emph{i.e.}, answers are open-ended) such as instruction following and multi-turn dialogue are predominantly assessed in English~\citep{alpaca_eval,zheng2023judging}. Existing multilingual extensions often rely on automated machine translation (MT) or cover only a handful of languages~\citep{zhang2024plug,liu2024omgeval}. This reliance on MT introduces “translationese” and fails to reflect native tongues, which can be trivial in objective tasks (\emph{e.g.}, true/false) but is especially harmful for subjective evaluation.

\textbf{(3) Lack of In-Depth Diagnostic Analysis.} Existing studies often stop at superficial leaderboard rankings~\cite{pomerenke2025ai,liu2024omgeval}, failing to reveal implications under performance variance. There is a scarcity of insights regarding how performance correlates with geographical regions, task types, or model architectures, leaving potential challenges and gaps untouched.

To address these challenges, we introduce \textbf{GaoYao}, a multilingual and multicultural benchmark emphasizing systematicity, authenticity, and analytical depth, which features three aspects:

\begin{figure*}[t]
    \centering
    \includegraphics[width=1\linewidth]{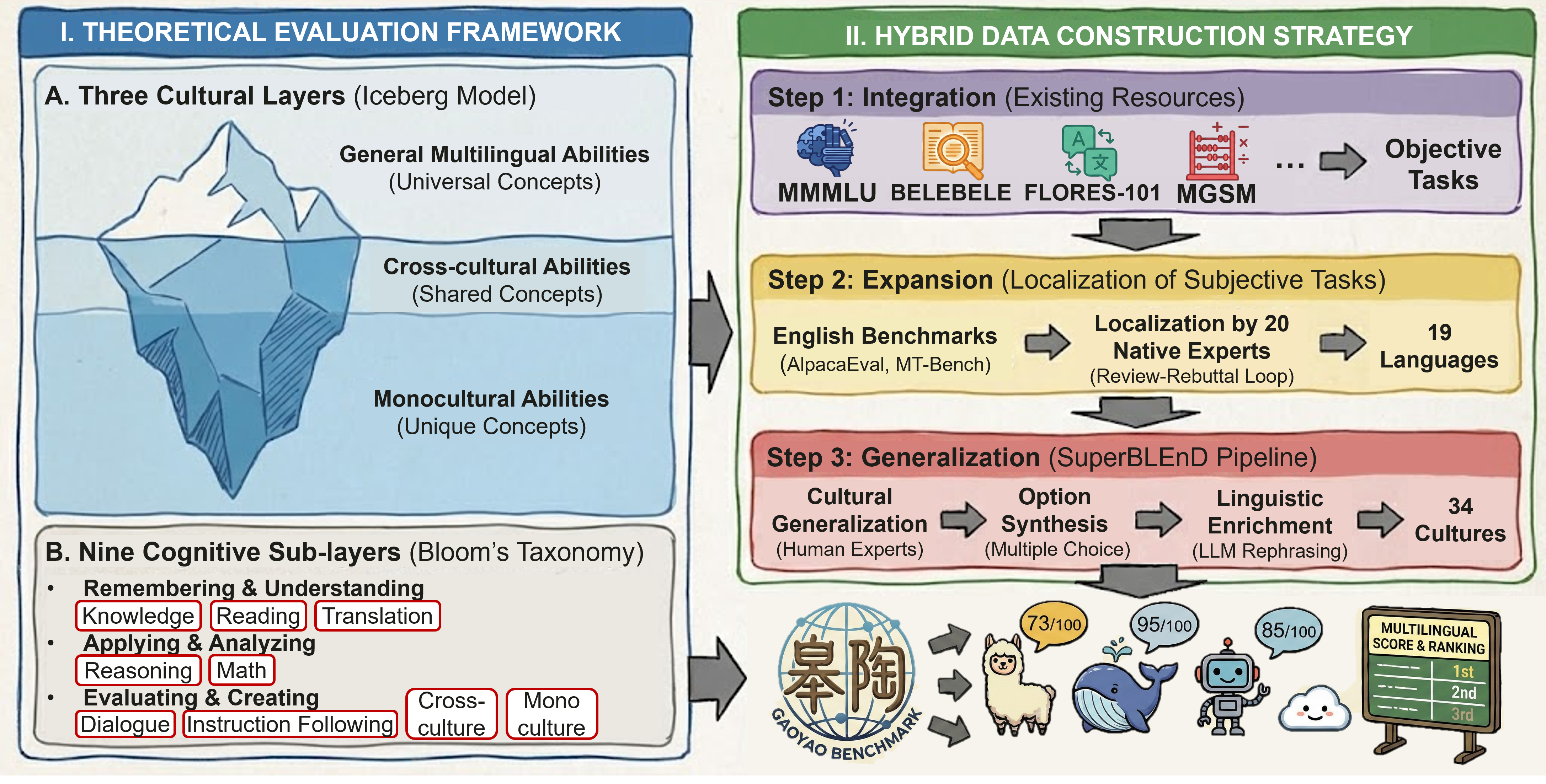}
    \caption{Illustration on design and construction of GaoYao. The benchmark is grounded in theoretical models of culture and cognition, and constructed through a hybrid strategy of integration, expansion and generalization.}
    \label{fig:main_methodology}
\end{figure*}

\textbf{(1) A Systematic Evaluation Framework.} Grounded in cultural theory~\citep{hall1976beyond} and cognitive taxonomy~\citep{anderson2001taxonomy}, we propose a unified evaluation matrix. This framework categorizes capabilities into three layers: \textit{General Multilingual Abilities} (universal concepts), \textit{Cross-cultural Abilities} (culturally shared concepts with variance), and \textit{Monocultural Abilities} (unique concepts in cultures). These are further expanded into nine cognitive sub-layers, ranging from knowledge retention to creative writing.

\textbf{(2) Native-Quality Data Expansion.} We address the data scarcity by leveraging a team of multilingual experts to meticulously localize English evaluation sets in two critical sub-layers into 19 languages, ensuring native-level quality compared to machine-translated alternatives. Additionally, we generalize a cultural evaluation set to cover 34 cultures through a novel expert-verified synthesis pipeline. In Fig.~\ref{fig:task_distribution_boxplot}, our three curated test sets are able to better reflect capability stratification of LLMs due to the native-level data quality.

\textbf{(3) In-Depth Diagnostic Findings.} We conduct a tiered evaluation of representative SOTA models and go beyond simple rankings. Our analysis reveals the severe "digital divide" across regions and the performance gap between mature and frontier tasks. Synthesizing these findings, we propose a \textit{meta-finding} to guide the community: recommending strategic deployment methods for efficient model usage and advocating for equitable data construction to bridge the capability gaps. 

Our contributions can be summarized as:
\begin{itemize}
    \item We propose a systematic evaluation benchmark with 23.3M tokens, three cultural layers and nine cognitive sub-layers, addressing the fragmented nature of existing benchmarks.
    \item We expand critical instruction-following and dialogue evaluations to 19 languages and generalize cultural evaluation sets to 34 cultures via a rigorous human-in-the-loop method, filling the blank with high data quality and enabling better identification of LLM abilities.
    \item We provide a comprehensive capability landscape of existing multilingual LLMs through deep analysis, revealing several key insights which guide LLM usage and development.
\end{itemize}

In addition, we release all test sets and evaluation codes, providing the community with a reliable compass for future work on multilingual LLMs.



\section{Methodology}
\label{sec:method}

As shown in Fig.~\ref{fig:main_methodology}, our framework begins by defining a theoretical landscape of multilingual and multicultural capabilities critical in LLM evaluation. Guided by this theoretical structure, we employ a three-pronged approach—Integration, Expansion, and Generalization—to curate a comprehensive benchmark that addresses existing gaps in systematicity, coverage and quality. Section~\ref{subsec:dimensions} details the theoretical underpinnings of our evaluation dimensions and Section~\ref{subsec:construction} discusses the specific processes involved in constructing the GaoYao benchmark through these three strategies.

\subsection{Layered Evaluation Dimensions} \label{subsec:dimensions}

\paragraph{Theoretical Foundations of Three Major Layers} Drawing inspiration from the Cultural Iceberg Model~\citep{hall1976beyond} and the Three-Layer Model of organizational culture~\citep{schein2010organizational}, we posit that existing multilingual benchmarks often predominantly assess “surface-level” linguistic proficiencies while overlooking the deeper, implicit cultural contexts that shape communication. To address this, GaoYao categorizes tasks into three major layers representing cultural deepening levels:

\begin{itemize} 
\item \textbf{General Multilingual Abilities:} This layer corresponds to the “tip of the iceberg,” focusing on universal concepts that remain consistent across languages (\emph{e.g.}, applying target language to handle problems involving reasoning, knowledge or comprehension). 

\item \textbf{Cross-cultural Abilities:} Moving beneath the surface, this layer assesses the model's capacity to navigate shared concepts that manifest differently across cultures. For instance, while the lexical term “dragon” translates directly, its symbolic meaning varies drastically: Western dragons are typically depicted as malevolent monsters while the eastern dragon (or loong) is revered as an auspicious symbol~\citep{zhao1988study}. An LLM must discern these subtle cultural divergences.

\item \textbf{Monocultural Abilities:} The deepest layer evaluates the understanding of unique concepts exclusive to specific cultures, which often lack direct equivalents elsewhere. An example is the Chinese phenomenon of “Chunyun” (the massive Spring Festival travel rush~\citep{zhu2021exploring}), a culturally specific event laden with unique social implications. Another example is “Namaste”, the special greeting etiquette in India~\citep{zhang2025culturescope}. \end{itemize}

\paragraph{Deriving Nine Sub-layers via Cognitive Taxonomy} Within these major cultural layers, we further ensure a comprehensive evaluation matrix by structuring tasks according to Bloom's Taxonomy of cognitive domains~\citep{anderson2001taxonomy}. This taxonomy categorizes human thought processes into six categories along a gradient of complexity, ranging from basic remembering to complex creation. Inspired by the six cognitive levels, the task design of GaoYao encompasses nine distinct sub-layers to ensure a rigorous assessment:

\begin{itemize} 
\item \textbf{Remembering \& Understanding:} Reflected by tasks of multilingual \textit{Knowledge} Q\&A, \textit{Reading} Comprehension, and \textit{Translation}. 

\item \textbf{Applying \& Analyzing:} Assessed through \textit{Reasoning} tasks and \textit{Math} problem solving.

\item \textbf{Evaluating \& Creating:} This highest cognitive level encompasses: (1) Creative Tasks: \textit{Instruction Following} and \textit{Multi-turn Dialogue}, which demand creative writing to satisfy complex, open-ended user intents; (2) Evaluative Tasks: The advanced \textit{Cross-cultural} and \textit{Monocultural} assessments. Unlike simple factual retrieval, these tasks require the model to evaluate social nuances, discern cultural appropriateness among highly plausible distractors, and make value judgments aligned with cultural norms~\cite{rystrom2025multilingual}.
\end{itemize}

\subsection{Construction of GaoYao Benchmark} \label{subsec:construction}

Guided by the theoretical framework above, we construct the GaoYao benchmark through a hybrid strategy combining the integration of established resources (for seven of the nine sub-layers), the linguistic expansion of high-value under-served benchmarks (two most critical sub-layers: instruction following and multi-turn dialogues), and the generalization of cultural data through human-in-loop synthesis pipelines (the cross-cultural layer).

\subsubsection{Integration of Existing Test Sets} \label{sec:integration}

For several of the defined cognitive sub-layers, particularly those related to objective knowledge and reasoning, the research community has already established high-quality open-source benchmarks. Rather than reinventing these, we conducted literature review and quality checks to select and integrate some of the most widely-verified and robust datasets into GaoYao, ensuring a complete coverage of our defined evaluation sub-layers. These datasets are mapped to our sub-layers as follows:

(1) \textit{Knowledge \& Reasoning}: Given their coverage on factual knowledge spanning various subjects from elementary-level knowledge up to advanced professional subjects, both \textsc{Include}~\citep{romanouinclude} and \textsc{MMMLU}~\citep{openai_mmmlu_2024} are integrated. Compared with \textsc{Include}, \textsc{MMMLU} focuses more on reasoning abilities, \emph{i.e.}, how LLMs apply these knowledge to solve practical problems. 

(2) \textit{Reading}: We incorporate \textsc{Belebele}~\citep{bandarkar-etal-2024-belebele} for evaluating multilingual reading comprehension capabilities given its native passage coverage and rigorous quality assurance. 

(3) \textit{Translation}: \textsc{Flores-101}~\citep{goyal2022flores} provides a widely-recognized standard for assessing MT across numerous language pairs. 

(4) \textit{Math}: \textsc{MGSM}~\citep{shilanguage} is also a widely-used dataset to evaluate multilingual mathematical reasoning capabilities. 

(5) \textit{Cross-culture \& Monoculture}: Since the research community has only recently begun to rigorously define and evaluate the multicultural capabilities of LLMs~\citep{rystrom2025multilingual}, open-source resources remain scarce. We leverage two recent datasets: \textsc{SAGE}~\citep{guo2025largelanguagemodelstruly} for identifying cultural differences in shared concepts (cross-culture) and \textsc{CultureScope}~\citep{zhang2025culturescope} for understanding unique cultural concepts (monoculture). While both datasets delve deeply into culture-specific concepts and employ rigorous design procedures, their coverage is restricted to Chinese and Spanish. To supplement this, we constructed a cross-cultural evaluation set spanning 34 cultures (see Section~\ref{sec:generalization}).

\subsubsection{Expansion of Language Coverage for \textsc{AlpacaEval} and \textsc{MT-Bench}} \label{sec:expansion}

Instruction following and multi-turn dialogue represent critical capabilities reflecting an LLM's practical utility and “human-likeness.” However, existing multilingual benchmarks heavily prioritize objective tasks, leaving these subjective, open-ended abilities predominantly evaluated only in English. To close this significant gap, we selected two widely recognized English benchmarks: \textsc{AlpacaEval}~\cite{alpaca_eval}, validated by over 20k human judgments for general instruction following, and \textsc{MT-Bench}~\cite{zheng2023judging}, designed with challenging multi-turn questions across intent categories such as role playing and creative writing. We then expanded their coverage to over 19 languages, denoting as \textsc{S-AlpacaEval} and \textsc{S-MT-Bench}, respectively.

This expansion was not a simple translation task but a rigorous localization effort. From the language service center of a top-tier corporation, we recruited a team of 20 native-speaker professionals with expertise in translation, localization, and linguistic testing. The team dedicated a total of 175 person-days to this development. To ensure the highest quality, a strict review-rebuttal feedback loop was implemented for each language. Third-party reviewers continuously inspected samples during annotation. Disagreements triggered a discussion phase where annotators either revised their work based on the concerns or provided justifications to persuade the reviewer to unflag the sample. 

Crucially, there is a localization process to make sure every user question is linguistically feasible, which can hardly be guaranteed using MT. For instance, constrained English instruction like “list items starting with the letter A” will be invalid if being translated literally to a language without letter A. Thus, such instructions were manually adapted or reconstructed to suit the phonetic and script characteristics of the target language while ensuring the cognitive task remained equivalent.

\subsubsection{Generalization of Cross-cultural Evaluation (\textsc{SuperBLEnD})} \label{sec:generalization}

As discussed in Section~\ref{sec:integration}, existing cultural evaluation sets are limited in its culture coverage. However, expanding such coverage presents a unique challenge: direct translation retains source-culture concepts, while manual creation is costly. To address this, we generalized \textsc{BLEnD}~\citep{myung2024blend} into \textsc{SuperBLEnD}, expanding coverage from 16 to 34 cultures via a three-stage semi-automated pipeline incorporating rigorous human verification. \textsc{SuperBLEnD} focuses on evaluating understanding of cultural differences regarding everyday concepts such as festivals, food and sports. The pipelines are as follows (full annotations and technical details are in Appendix~\ref{sec: detailed superblend annotation}):

\begin{figure}[bp]
    \centering
    \includegraphics[width=\linewidth]{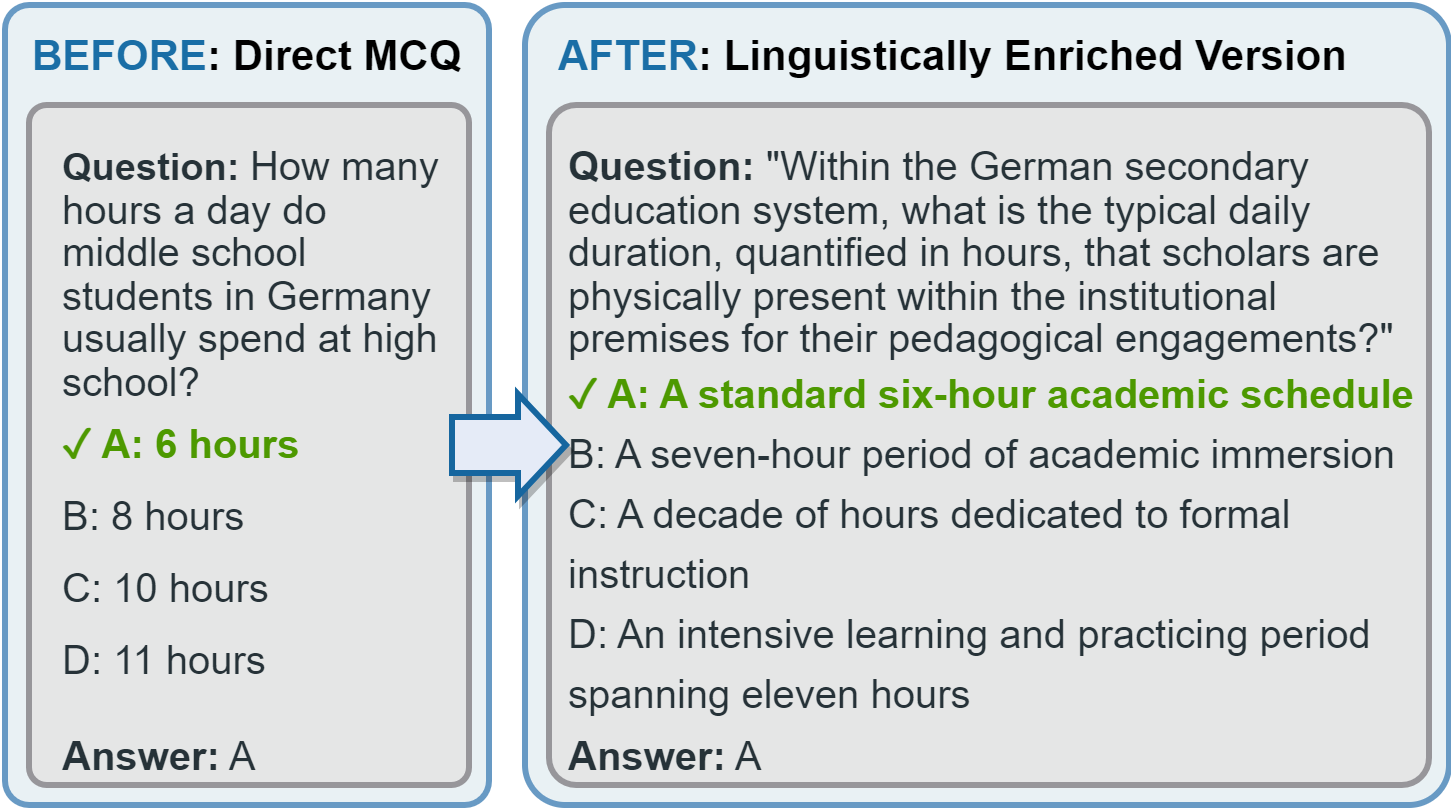}
    \caption{An example of the linguistic enrichment process (stage 3), which increases complexity of MCQs without altering the underlying cultural fact.}
    \label{fig:rephrasing_example}
\end{figure}

\paragraph{Stage 1: Cultural Generalization.} We curated a subset of high-quality templates from \textsc{BLEnD} (inheriting answers for the original 16 cultures) and recruited native experts to provide authentic answers for 18 additional cultures based on lived experience. Answers underwent strict manual verification to eliminate invalid or toxic content (discarding $\sim$41.1\% of raw data), ensuring high-quality ground truth even for questions with multiple valid answers (\emph{e.g.}, accepting both "beer" and "carbonated drinks" for Malaysian nightclubs).

\paragraph{Stage 2: Option Synthesis.} To enhance diversity, verified Q\&A pairs from stage 1 were converted into multiple-choice questions (MCQs) by combining target answers with distractors from other cultures or plausible LLM-generated "dummy options". Options underwent strict verification to exclude low-quality cases such as hierarchical conflicts (\emph{e.g.}, rejecting "Pepsi" as a distractor if the answer is "beer", as it falls under the valid category of "carbonated drinks").

\paragraph{Stage 3: Linguistic Enrichment.} To enhance difficulty and prevent simple pattern matching, we utilized an LLM to rephrase question stems and options via techniques like syntactic restructuring and voice alternation. As shown in Fig.~\ref{fig:rephrasing_example}, this process ensures the benchmark tests deep cultural reasoning rather than superficial keyword recognition.

\section{Experiment}
\label{sec:exp}

\subsection{Experimental Setups}
To empirically validate GaoYao's efficacy in mapping the global LLM landscape, we conducted a tiered evaluation across a spectrum of models representing the current SOTA. Our selection encompasses both open-weights models (\emph{e.g.}, DeepSeek-V3.1~\cite{deepseek2025v31}) inferred on standardized NPU computation nodes, and proprietary commercial models (\emph{e.g.}, GPT-5~\cite{openai2025gpt5}) accessed via official APIs. The specific model versions and resource addresses are in Table~\ref{tab: model addresses}. We make sure all evaluated LLMs are post-trained versions (\emph{e.g.}, “instruct” or “chat” versions) with “thinking” disabled (except Fig.~\ref{fig:thinking_impact}). Following \citet{yang2025qwen3}, we adopted only a random 10\% subset of \textsc{MMMLU} due to its unproportionate volume. Section~\ref{sec:Statistics of GaoYao} and Section~\ref{sec:evaluation approaches} further illustrates the setups. See a reliability analysis of GaoYao in Appendix~\ref{sec:reliability}.

\subsubsection{Statistics of Test Sets in GaoYao}\label{sec:Statistics of GaoYao}
\begin{figure}
    \centering
    \includegraphics[width=\linewidth]{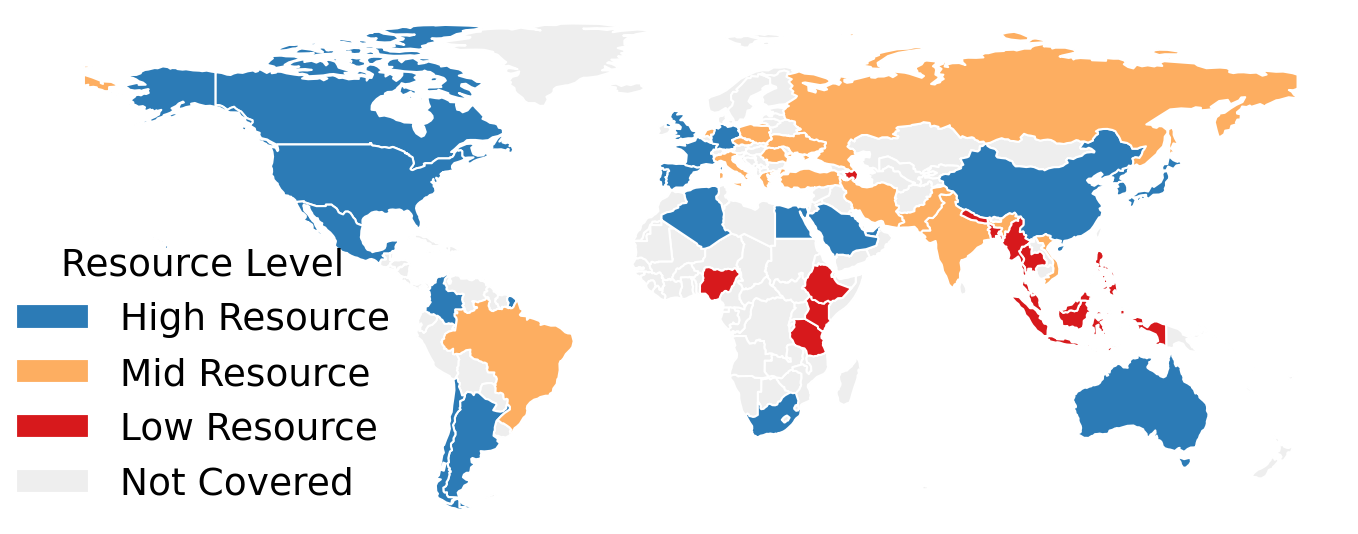}
    \caption{The language and culture coverage on the world map. Colors indicate resource popularity levels.}
    \label{fig:gaoyao_coverage_map}
\end{figure}

\begin{figure}[tbp] 
\centering
\begin{minipage}{0.95\linewidth}
  \centering
  \includegraphics[width=\linewidth]{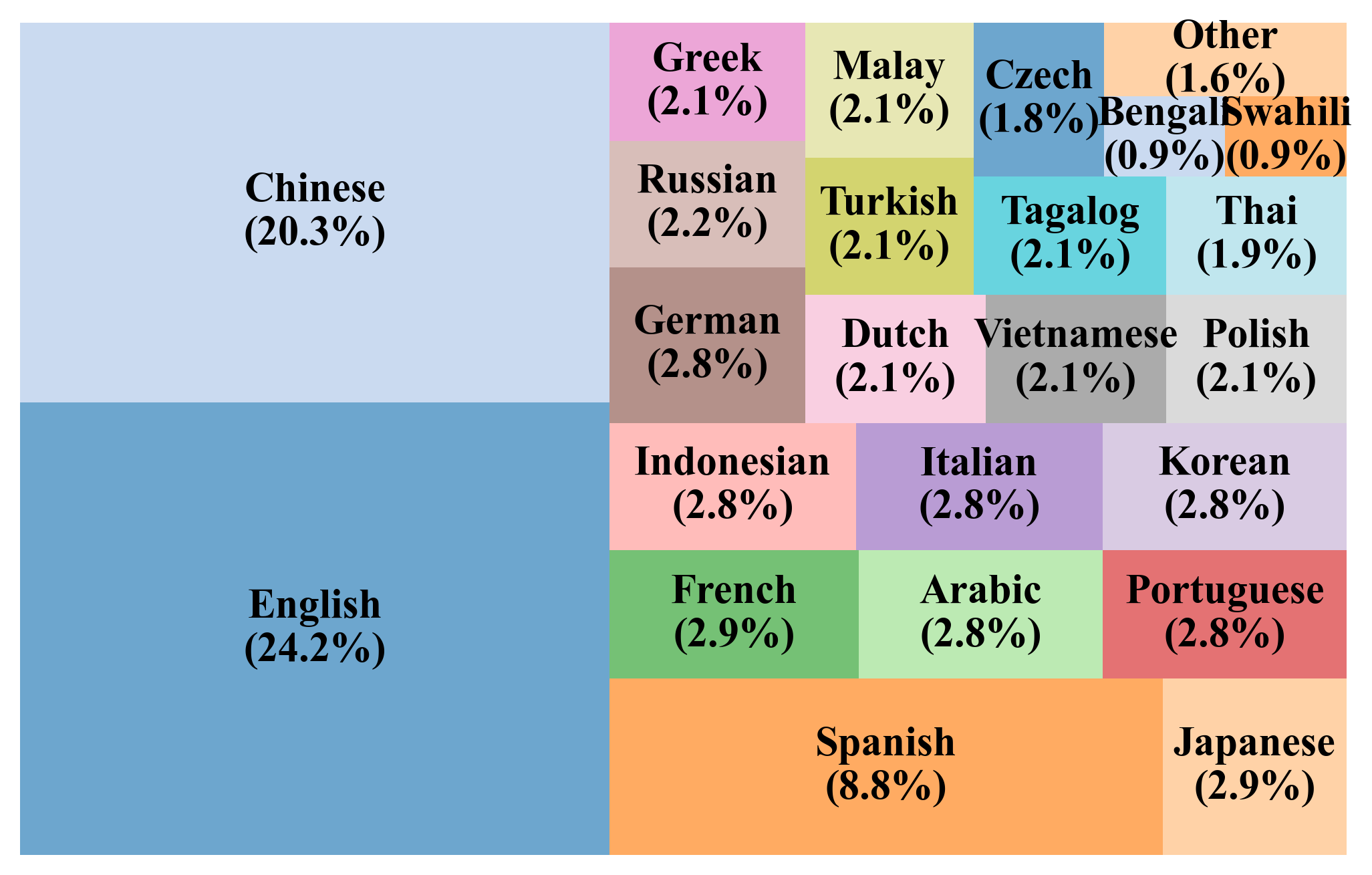}\\
  (a) Sample distribution by language
\end{minipage}
\hfill
\begin{minipage}{\linewidth}
  \centering
  \includegraphics[width=\linewidth]{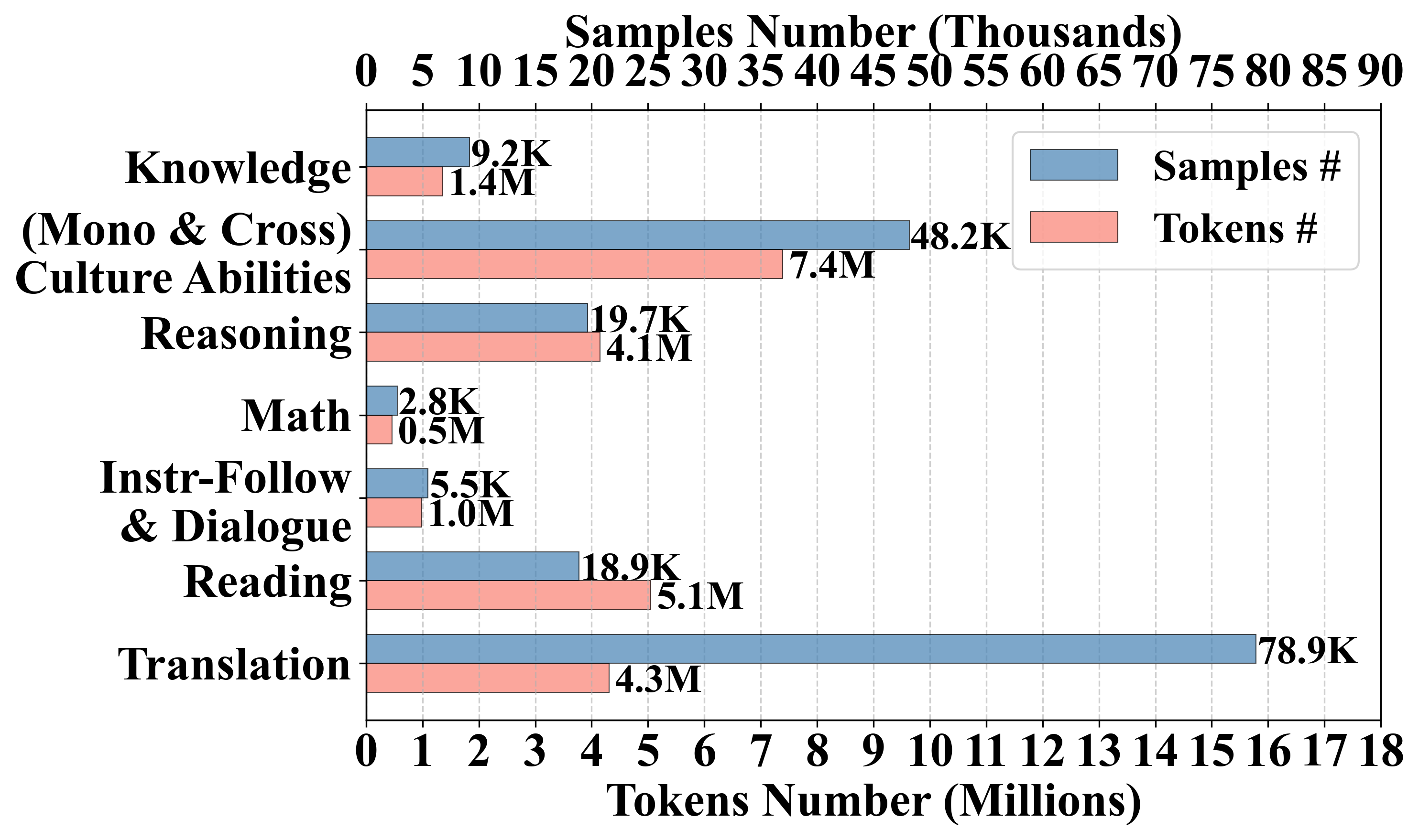}\\
  (b) Distribution by evaluation dimensions
\end{minipage}
\caption{Distribution statistics of test sets in GaoYao (a) by languages and (b) by evaluation sub-layers.}
\label{fig:dataset_distribution} 
\end{figure}

\begin{figure*}[t!]
 \centering  
 \subfigbottomskip=-2pt 
 \subfigcapskip=-2pt 
 \subfigure[Open-Source Models]{
  \includegraphics[width=0.85\linewidth]{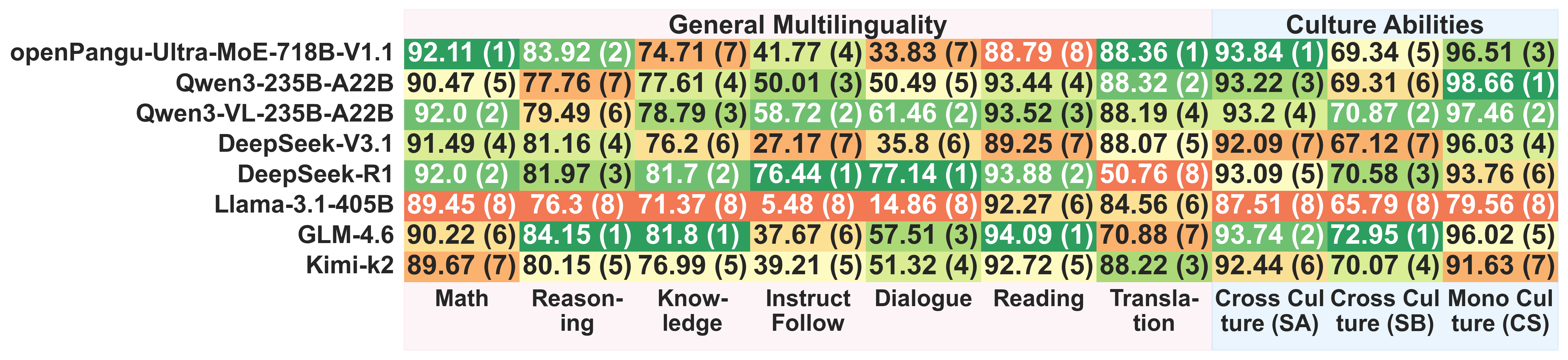}}
 \subfigure[Closed-Source (API-based) Models]{
  \includegraphics[width=0.85\linewidth]{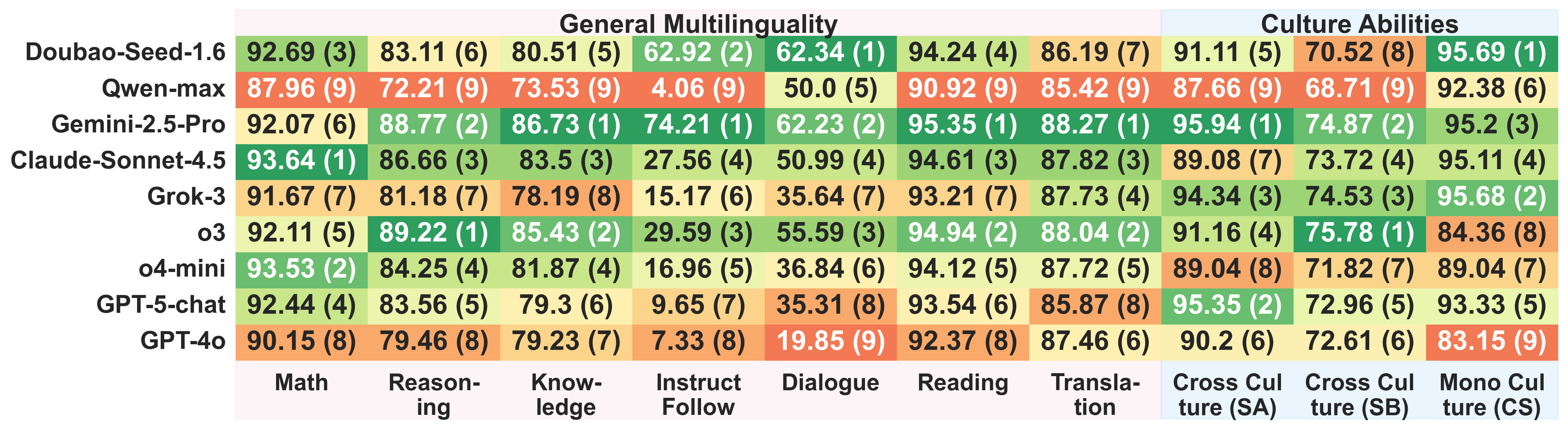}}
 \subfigure[Compact Models ($<20$B)]{
  \includegraphics[width=0.85\linewidth]{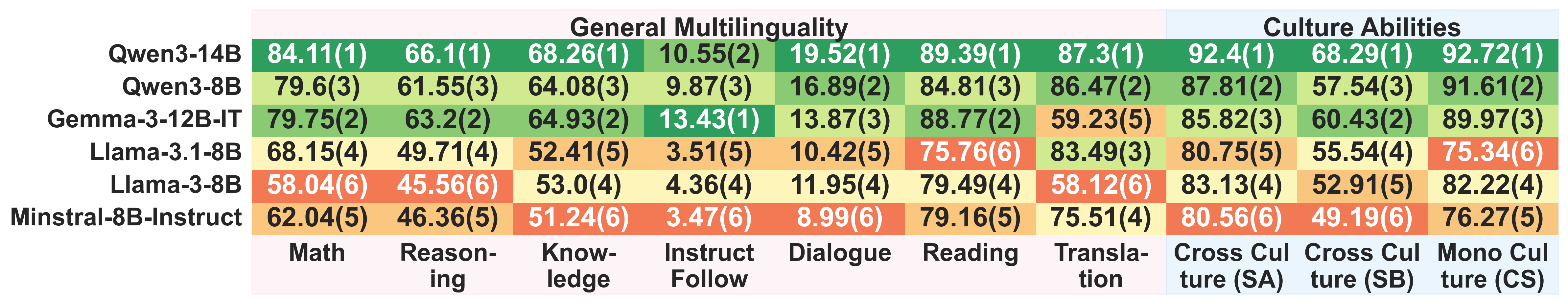}}
 \caption{Performance heatmaps across nine evaluation sub-layers. Scores are averaged across all languages. Numbers in parentheses indicate rank within the group. Backgrounds: Pink (General Multilingual), Blue (Cultural Abilities). SA, SB and CS represents specific datasets: \textsc{SAGE}, \textsc{SuperBLEnD} and \textsc{CultureScope}.}
\label{fig:Scores_distribution_per_dataset}
\end{figure*}

As shown in Fig.~\ref{fig:gaoyao_coverage_map}, the dataset spans 26 languages distributed across 51 nations/areas (34 of them are culturally represented as discussed in Section~\ref{sec:generalization}). The distribution encompasses five geopolitical clusters: Western Europe, Eastern Europe, East Asia \& Southeast Asia, Middle East \& Africa, and South Asia (See Table~\ref{tab:languages_table} for detailed statistics). A crucial design principle of GaoYao is the mitigation of resource bias; as illustrated in Fig.~\ref{fig:dataset_distribution}(a), excluding the three dominant lingua francas, the distribution is relatively balanced, with each remaining language constituting roughly 1\%-3\% of the total volume in sample level. According to \citet{joshi2020state}, the languages include nine low-resource and ten mid-resource varieties (See Appendix~\ref{sec:language_detail_info}).

Fig.~\ref{fig:dataset_distribution}(b) shows the distribution of test set sizes across the nine evaluation sub-layers of the GaoYao benchmark, as introduced in Section~\ref{subsec:dimensions}. The distribution of samples and tokens varies distinctively across sub-layers due to their inherent task characteristics. While \textit{Translation} comprises the highest volume of samples, it accounts for a relatively modest share of total tokens, reflecting the sentence-level brevity typical of the \textsc{Flores-101} dataset. In contrast, sub-layers such as \textit{Reasoning} and \textit{Reading} exhibit a significantly higher token-to-sample ratio, as these domains necessitate extensive context to define complex problem spaces. Additionally, the cultural layers (\textit{Monoculture} and \textit{Cross-culture}) represent a relatively large token count to ensure sufficient depth to capture cultural nuances, reflecting GaoYao’s emphasis on cultural evaluation.

\subsubsection{Evaluation Approaches}~\label{sec:evaluation approaches}
The evaluation protocol for each sub-dataset can be divided into two categories (details on metrics, judges and calculation methods are in Table~\ref{tab:evaluation_method}):

\textbf{Objective Evaluation:} For question type with deterministic outputs (\emph{e.g.}, MCQ, calculation problems), we utilize standardized prompt templates~\citep{openai_mmmlu_2024, romanouinclude, bandarkar-etal-2024-belebele, goyal2022flores} and rule-based extraction with regular expressions to parse answers from LLMs' responses. To ensure reproducibility, all pre-processing and post-processing scripts have been released.

\textbf{Subjective Evaluation:} For open-ended tasks (\emph{e.g.}, Q\&A), we adopt the widely-used “LLM-as-Judge" paradigm~\citep{alpaca_eval, zheng2023judging}, where the judge model compares response from a candidate model with a reference response based on specific dimensions and concludes with “win”, “lose” or “tie”. We standardized on DeepSeek-v3.1 as the judge due to its superior reasoning abilities. The primary metric is \textit{Win Rate} against the reference responses. For datasets lacking inherent references (\textsc{S-AlpacaEval}, \textsc{S-MT-Bench}), we introduced Qwen3-235B-A22B~\cite{yang2025qwen3} as the reference anchor. 

All scores (\emph{e.g.}, accuracy, win rate) are displayed at the scale of 0-100 for clearer viewing. The results are aggregated along specific axes, \emph{e.g.}, \textit{Task Dimension} (averaging across all languages for a specific evaluation sub-layer) for Finding 1\&3 and \textit{Language Dimension} (averaging across all sub-layers for a specific language) for Finding 2.

\subsection{Results and Findings}

\paragraph{Finding 1: Performance Differentiation Among Flagship and Compact Models} \label{finding 1}

Fig.~\ref{fig:Scores_distribution_per_dataset}(a) and (b) present the landscape of flagship models, including open-source leaders and closed-source commercial APIs. The results reveal distinct multilingual capability profiles rather than a uniform dominance:

\begin{itemize}
    \item \textit{Logic \& Culture Specialist:} openPangu-Ultra-MoE-718B-V1.1~\cite{openpangu2024ultra} exhibits exceptional strength in \textit{Math} (\#1) and \textit{Reasoning} (\#2) among open-source LLMs, while simultaneously securing among top ranks in the \textit{Cross-culture} sub-layer. This correlation suggests its rigorous logical training may facilitate understanding of complex cultural frameworks.
    \item \textit{Knowledge Heavyweights:} Gemini-2.5-Pro \cite{comanici2025gemini} demonstrates dominance in \textit{Knowledge}, \textit{Reading} and \textit{Translation}, suggesting a pre-training corpus with extensive informational breadth.
    \item \textit{Interaction Specialists:} DeepSeek-R1 \cite{deepseek2025r1} leads both open-source and closed-source models in \textit{Instruction Following} and \textit{Dialogue}, reflecting a post-training strategy optimized for conversational utility and complex user constraints.
\end{itemize}

\begin{figure}[tbp]
  \centering
  \includegraphics[width=\linewidth]{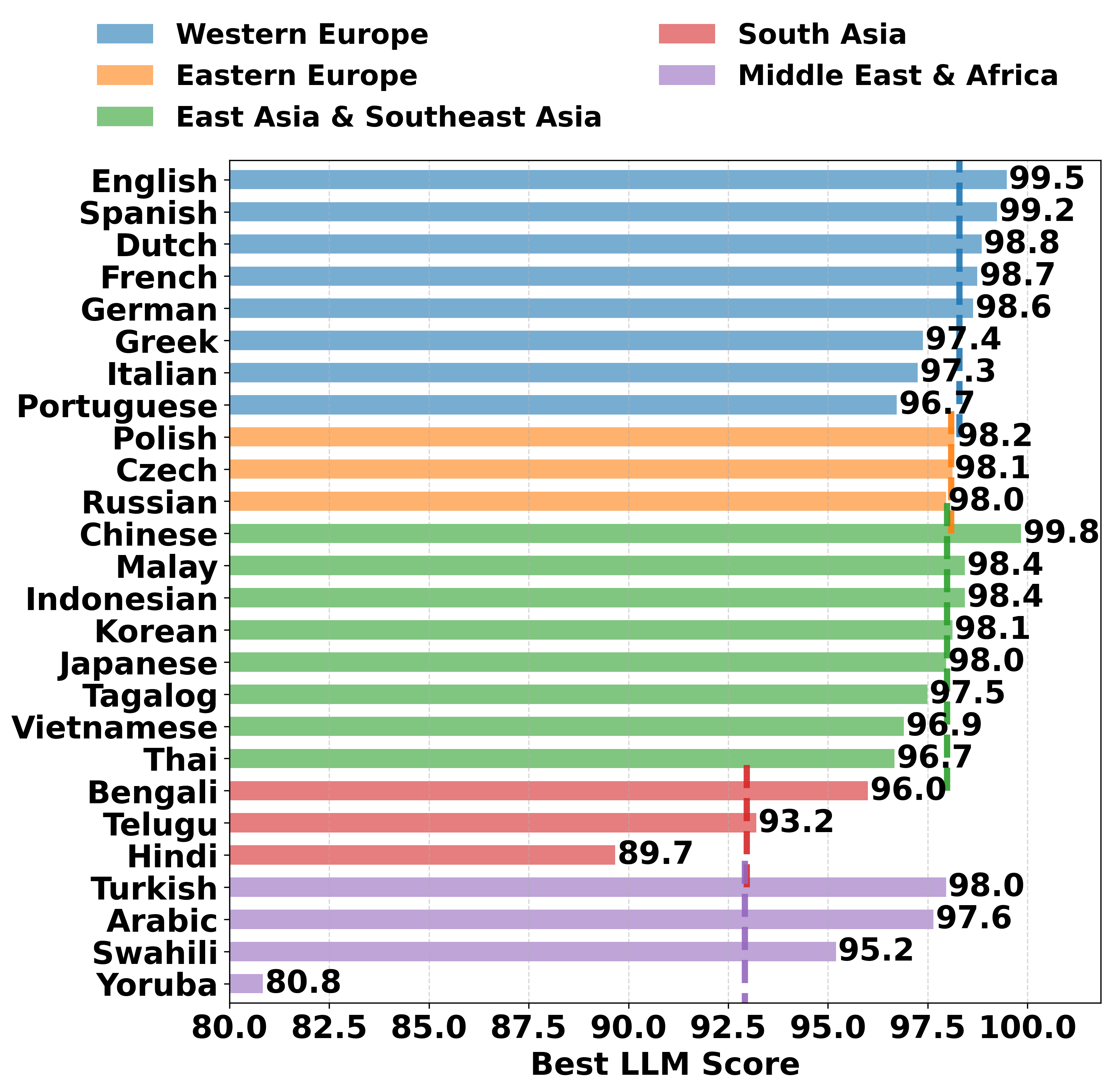}
\caption{Impact of geography on \textbf{best} performance achieved by LLMs across languages. Vertical dashed lines represent group averages.}
\label{fig:langage_distribution_scores}
\end{figure}

Fig.~\ref{fig:Scores_distribution_per_dataset}(c) illustrates the performance of compact models ($<20$B parameters). We observed a possibility of saturation for open-source benchmark: A crucial finding is the discrepancy in performance gaps. On established benchmarks like \textsc{Belebele} and \textsc{Include}, the compact Qwen3-14B~\cite{yang2025qwen3} achieves near-parity with the massive Qwen3-235B. However, on GaoYao's newly constructed subjective sets (\textsc{S-AlpacaEval} and \textsc{S-MT-Bench}), a significant gap persists. This suggests that some popular open-source benchmarks may have become insensitive in identifying multilingual capabilities (a phenomenon called benchmark saturation)~\cite{akhtar2026ai}, whereas GaoYao's fresh, expert-localized data exposes the true gap between compact and flagship models. Finding 3 further investigates it.

\paragraph{Finding 2: Digital Divide Among Language Geography and Resource-Levels} \label{finding 2}
As shown in Fig.~\ref{fig:langage_distribution_scores} and Fig.~\ref{fig:langage_distribution_scores_popularity}, by analyzing LLMs' best performances through a geopolitical and resource-level lens (\emph{i.e.}, aggregating highest scores among all LLMs by languages), a persistent "digital divide" is revealed. Performance on a certain language is strongly correlated with geographic attributes and resource availability: Western European languages consistently score highest, while low-resource languages in South Asia and Africa lag significantly. This hierarchy is consistent with resource levels in Fig.~\ref{fig:langage_distribution_scores_popularity}: High > Medium > Low popularity across maximum, mean, and minimum scores, underscoring that current multilingual progress is uneven and largely driven by data volume rather than universal linguistic transfer.

\begin{figure}[t]
  \centering
  \includegraphics[width=\linewidth]{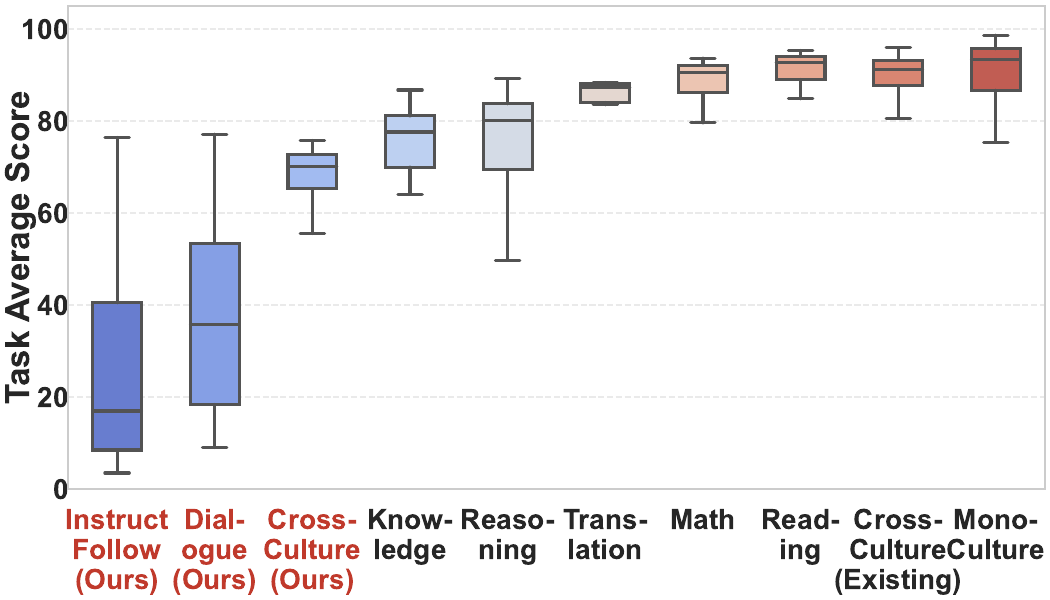}
  \caption{Distribution of model performance across tasks (\emph{i.e.}, sub-layers) using standard box plot. Height of boxes and whiskers indicate the performance divergence among models, while the horizontal line is the median score. Datasets constructed in this work are in \textbf{\textcolor[HTML]{C0392B}{Red}}.}
  \label{fig:task_distribution_boxplot}
\end{figure}

\paragraph{Finding 3: Capability Stratification Between Solved and Frontier Tasks} \label{finding 3}

Fig.~\ref{fig:task_distribution_boxplot} presents a boxplot analysis of model performance distributions, revealing a severe stratification in current LLM capabilities for different tasks (\emph{i.e.}, sub-layers). Objective tasks such as \textit{Reading} and \textit{Math} exhibit high median scores ($>85$) with compressed boxes, indicating a closed gap between flagship and compact models due to their standardized patterns. In contrast, subjective tasks (especially for \textit{Instruction Following} and \textit{Dialogue} with expert-localized datasets in GaoYao) display significantly lower medians and elongated interquartile ranges (box and whisker heights). This high variance confirms that these tasks serve as high-sensitivity discriminators due to their advanced requirements on creating and human-likeness, effectively exposing the capability frontier where flagship models significantly outperform average open-source models.

In addition, a comparison within cultural tasks demonstrates the value of the \textsc{SuperBLEnD} dataset. While the two existing cultural test sets (\emph{i.e.}, \textsc{SAGE} and \textsc{CultureScope}) show signs of saturation (medians $\approx 90$), our synthesized dataset for the cross-cultural layer reveals a significant drop in median score and relatively wide dispersion. This proves that the construction strategy in Section~\ref{sec:generalization} successfully elevates the challenge from simple knowledge retrieval to rigorous cultural reasoning on authentic experiences, leading to a more precise test set for cultural capabilities.

\begin{figure}[t]
  \centering
  \includegraphics[width=\linewidth]{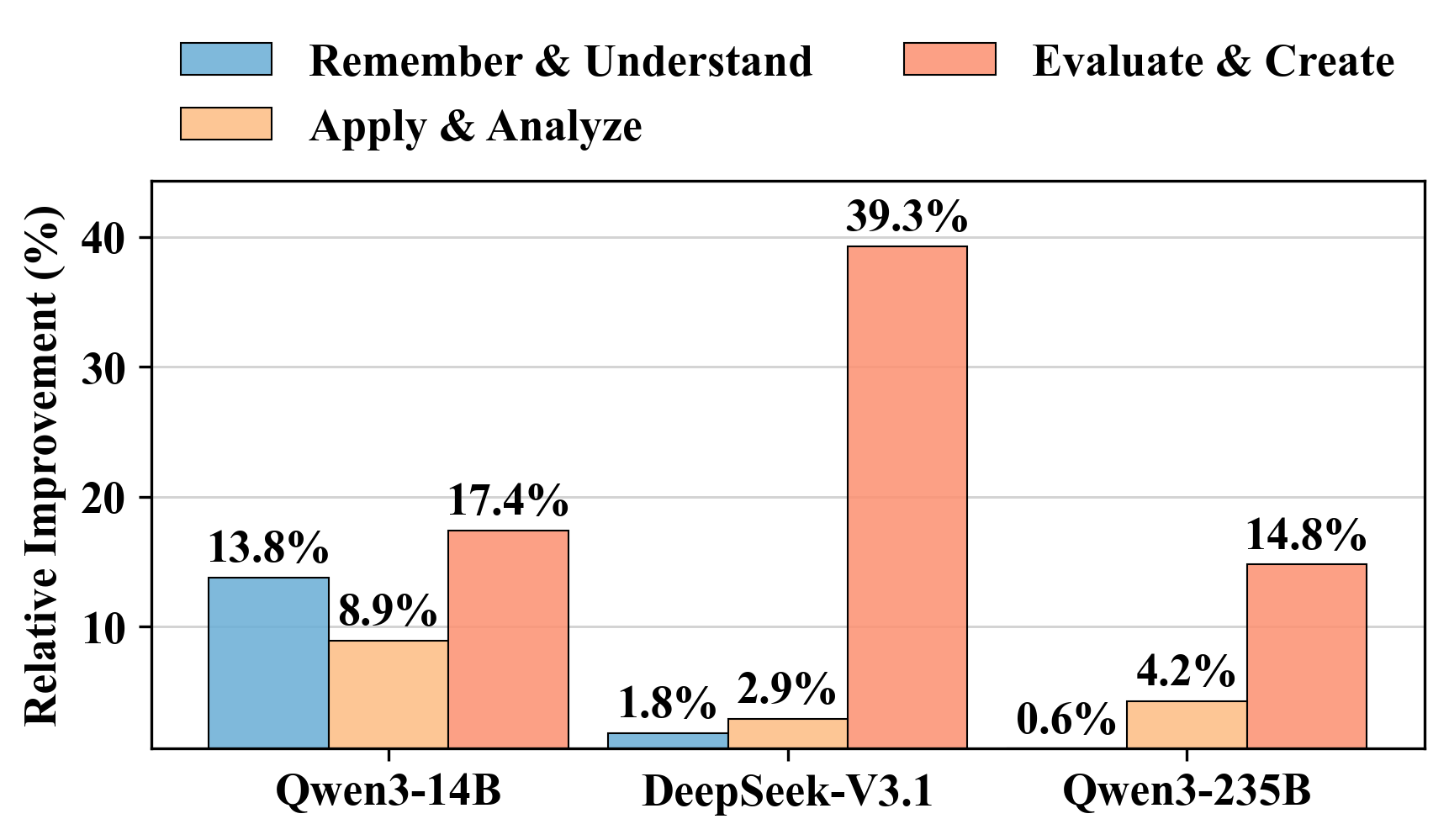}
  \caption{Multilingual performance gain by “thinking” mode ($\frac{\text{Think} - \text{Base}}{\text{Base}}$*100\%) across Bloom cognitive layers.}
  \label{fig:thinking_impact}
\end{figure}

\paragraph{Finding 4: Uneven Gain by “Thinking” in Multilingual Tasks} \label{finding 4}

The paradigm of inference-time reasoning ( or “thinking”) has been proven effective in massive fields~\cite{deepseek2025r1,liu2025r,he2025r1}. In Fig.~\ref{fig:thinking_impact}, we investigated the impact of “thinking” in multilingual context. By aggregating task scores across Bloom's cognitive taxonomy (as described in Section~\ref{subsec:dimensions}), two divergent behaviors are revealed:

(1) \textit{Selective Gain for Flagships.} For flagship models (DeepSeek-V3.1, Qwen3-235B), the "Thinking" mode acts as a specialized tool. In the \textit{Remembering \& Understanding} layer (\emph{e.g.}, Translation, Knowledge), gains are marginal, suggesting the bottleneck is mainly multilingual knowledge for retrieval-heavy tasks. However, in the \textit{Evaluating \& Creating} layer (\emph{e.g.}, Instruction Following), we observe significant gains, probably due to a more comprehensive considerations of the constraints in users' instructions.

(2) \textit{Universal Gain for Compact Models.} In contrast, the compact Qwen3-14B benefits universally, achieving significant gains even in basic understanding tasks. This suggests that “Thinking” effectively compensates for the limited parameter capacity of smaller models, allowing them to punch above their weight.

\paragraph{Meta-Finding: From Benchmarking to Guidance} \label{meta-finding}

Transcending individual metrics, our findings (F1-F4) combine into a meta-finding that guides multilingual LLM utility and development:

\textbf{Strategic Deployment (Usage).} There is no "one-fits-all" model. Users should choose flagship models wisely based on their features (the interaction specialist or the knowledge heavyweights, per F1) and adopt a dynamic strategy for “thinking” mode (F4): deploy compact models with “thinking” enabled for better performance in resource-constrained inference, while reserving flagship models for complex creative tasks to balance cost and performance.

\textbf{Equitable Construction (Development).} The geographic performance cliffs (F2) and task stratification (F3) reveal the training data gap both for low-resource areas and highly subjective tasks. Future data development must pivot from English-centric translation to authentic regional curation. We recommend leveraging the \textit{human-in-the-loop generalization} pipeline (in Section~\ref{sec:generalization}) to efficiently fill the training data voids, ensuring both language equity and authenticity.

\section{Discussion}
\label{sec:discussion}

\subsection{Necessity of the Hierarchical Framework}
\label{sec:discussion_hierarchy}

To further examine whether the three-layer framework reveals meaningful capability distinctions beyond an aggregation of existing benchmarks, we conduct a rank-based transfer analysis across the three major layers in GaoYao: General multilingual, Cross-cultural, and Monocultural abilities. Following the model set in Fig.~\ref{fig:Scores_distribution_per_dataset}, we rank the 23 evaluated models by their average scores within each layer and compute Spearman's rank correlation between the General multilingual ranking and the rankings of the two deeper cultural layers.

\begin{table}[t]
\centering
\small
\setlength{\tabcolsep}{4pt}
\resizebox{\linewidth}{!}{
\begin{tabular}{lccc}
\toprule
\textbf{Model} & \textbf{General} & \textbf{Cross-cultural} & \textbf{Monocultural} \\
\midrule
Gemini-2.5-Pro & \#1 & \#1 & \#8 \\
Doubao-Seed-1.6 & \#2 & \#14 & \#6 \\
Qwen3-235B-A22B & \#9 & \#11 & \#1 \\
DeepSeek-V3.1 & \#15 & \#16 & \#4 \\
\bottomrule
\end{tabular}
}
\caption{Representative model rankings across the three major layers in GaoYao. Higher general multilingual ranking does not necessarily transfer to deeper cultural layers.}
\label{tab:hierarchy_rank_analysis}
\end{table}

The results show only modest transfer from general multilingual capability to deeper cultural capabilities: the correlation is higher for Cross-cultural tasks (Spearman's $\rho=0.74$) but drops for Monocultural tasks ($\rho=0.61$). This decline is important because monocultural evaluation requires models to recognize culturally unique concepts and norms rather than solve language-general problems. As shown in Table~\ref{tab:hierarchy_rank_analysis}, Gemini-2.5-Pro ranks first in the General Multilingual and Cross-cultural layers but drops to eighth in the Monocultural layer, while DeepSeek-V3.1 rises from fifteenth in General Multilingual to fourth in Monocultural. Similarly, Qwen3-235B-A22B ranks ninth in General Multilingual but first in Monocultural. These rank shifts indicate that strong surface-level fluency or general reasoning does not guarantee deep cultural nuance.

This analysis validates the necessity of GaoYao's hierarchical framework. If all tasks were collapsed into a single multilingual score, these capability decouplings would be obscured. By separating General Multilingual, Cross-cultural, and Monocultural layers, GaoYao can diagnose where a model's multilingual competence genuinely transfers and where culturally grounded evaluation remains a distinct frontier.

\subsection{Necessity of Linguistic Enrichment in \textsc{SuperBLEnD}}
\label{sec:discussion_superblend}

We also conduct an ablation study to verify whether the option synthesis and linguistic enrichment strategies in \textsc{SuperBLEnD} make the benchmark more discriminative. We compare model accuracy on the original \textsc{BLEnD} setting without our enrichment against \textsc{SuperBLEnD} on the same 16 overlapping cultures. This controls for culture coverage, so the main difference is whether our synthesized distractors and enriched phrasings are applied.

Table~\ref{tab:superblend_ablation} shows that all three representative models experience accuracy drops after applying our synthesis and enrichment strategies. The decrease is relatively modest for larger flagship models, such as Qwen3-235B (-4.51) and GPT-5-chat (-8.07), but significantly more substantial for the compact Qwen3-8B (-20.81). This pattern suggests that the enriched benchmark removes shortcuts based on surface forms and keyword associations, forcing models to resolve the underlying cultural context among plausible distractors.

The ablation also restores a more expected capability hierarchy. Without enrichment, Qwen3-8B unexpectedly outperforms Qwen3-235B-A22B on the original \textsc{BLEnD} subset (78.06 vs. 72.57), suggesting that the original format may permit shortcut exploitation. After enrichment, Qwen3-235B-A22B becomes clearly more robust than Qwen3-8B (68.06 vs. 57.25). Therefore, linguistic enrichment is not merely a stylistic transformation; it improves diagnostic validity by making \textsc{SuperBLEnD} better distinguish culturally grounded reasoning from shallow pattern matching.

\begin{table}[t]
\centering
\small
\setlength{\tabcolsep}{4pt}
\resizebox{0.92\linewidth}{!}{
\begin{tabular}{lccc}
\toprule
\textbf{Model} & \textbf{\textsc{BLEnD}} & \textbf{\textsc{SuperBLEnD}} & \textbf{$\Delta$} \\
\midrule
Qwen3-235B-A22B & 72.57 & 68.06 & \textbf{-4.51} \\
Qwen3-8B & 78.06 & 57.25 & \textbf{-20.81} \\
GPT-5-chat & 78.45 & 70.38 & \textbf{-8.07} \\
\bottomrule
\end{tabular}
}
\caption{Ablation of \textsc{SuperBLEnD} on the 16 cultures overlapping with \textsc{BLEnD}. Scores are average accuracy.}
\label{tab:superblend_ablation}
\end{table}

\section{Conclusion}
In this work, we introduced GaoYao, a holistic benchmark designed to map the full spectrum of multilingual intelligence. Unlike fragmented prior efforts, GaoYao establishes a unified framework covering three cultural layers and nine cognitive dimensions. Through a rigorous \textit{human-in-the-loop} construction pipeline, we addressed the critical scarcity of high-quality resources for subjective and cultural tasks, proving that authentic evaluation requires native expertise rather than automated translation. GaoYao stands as a robust alternative to English-centric evaluations, guiding the field to move beyond surface-level linguistic fluency towards deep cultural alignment and equitable global access. 

Future work include expanding coverage of domains (\emph{e.g.}, agent abilities) and languages, and developing a dynamic leaderboard to keep up with the latest iterations of models.

\clearpage
\section{Limitations}

Despite our comprehensive efforts, GaoYao has several limitations that outline future directions:

\textbf{(1) Domain and Task Coverage:} Currently, GaoYao focuses primarily on general multilingual and multicultural capabilities. We do not currently cover specialized vertical domains (\emph{e.g.}, legal, medical, financial) or agentic capabilities (\emph{e.g.}, tool use, API calling) in multilingual contexts. However, we argue that a robust general-purpose multilingual foundation is a prerequisite for these specialized abilities. Constructing high-quality benchmarks for such specific domains requires advanced expertise that falls outside the scope of this foundational work.
    
\textbf{(2) Static Nature of Benchmarking:} The landscape of LLMs evolves at an unprecedented pace, and a static publication inevitably lags behind the release of the very latest models. To address this, we have open-sourced all test sets and evaluation code, ensuring full transparency and enabling third-party model developers to easily verify their own systems against GaoYao. Furthermore, we plan to launch and maintain a dynamic online leaderboard to continuously track the community's progress.
    
\textbf{(3) Scalability of Human-in-the-loop Pipeline:} Our insistence on native expert curation and verification ensures high data quality but inherently limits scalability compared to fully automated pipelines. Expanding to hundreds of low-resource languages using this rigorous standard is resource-intensive. Yet, we believe that in the current era of ubiquitous machine-generated content, establishing a high-quality, human-verified gold standard for a representative set of languages is more critical than broad but low-quality coverage.

\textbf{(4) Task and Language Imbalance:} GaoYao prioritizes high-quality and expert-verified coverage over perfectly uniform coverage across all sub-layers. As a result, some integrated resources naturally differ in language scope; for example, \textsc{MGSM} covers 10 languages, while cultural resources such as \textsc{SAGE} and \textsc{CultureScope} are limited to 2 languages/cultures. This imbalance reflects the current scarcity of reliable multilingual and multicultural evaluation resources rather than an assumption that all languages are equally represented in every task. Future versions will expand under-covered task-language pairs while preserving the same native-expert verification standard.

\section{Ethical Considerations}

We reveal the following ethical considerations of GaoYao:

\textbf{(1) Benchmark Usage and Contamination:} We release GaoYao to facilitate the assessment of LLMs. We explicitly discourage the inclusion of our test sets into model training corpora (contamination), which would render the evaluation validity null. We urge the community to treat this benchmark as a diagnostic tool rather than a leaderboard to be gamed.

\textbf{(2) Annotator Fair Compensation and Well-being:} All data annotators and linguistic experts involved in the localization and \textsc{SuperBLEnD} synthesis processes were full-time employees of professional language service providers and participated as part of their regular duties. They received standard professional salaries above local minimum wage, were informed about the intended use of the data, and were not recruited through unpaid labor or low-paid crowdsourcing. No personally identifiable information was collected.
    
\textbf{(3) Mitigation of Cultural Stereotypes:} Constructing cultural benchmarks carries a risk of reinforcing stereotypes. To mitigate this, we implemented a careful review process where native experts explicitly screened for offensive content, harmful generalizations, or political sensitivity as specified in Appendix~\ref{sec: detailed superblend annotation}. While we strive for neutrality, we acknowledge that cultural data may still reflect the subjective perspectives of the annotators.

\bibliography{main}

\appendix

\section{Related Work}

\begin{table*}[t]
\centering
\small
\setlength{\tabcolsep}{3.5pt}
\resizebox{\linewidth}{!}{
\begin{tabular}{lcccccc}
\toprule
\textbf{Benchmark} & \textbf{Focus Domain} & \textbf{Task Diversity} & \textbf{\# Total Langs} & \textbf{\# Subj. Langs\textsuperscript{\dag}} & \textbf{\# Cultures} \\
\midrule
\textsc{Flores-101}~\citep{goyal2022flores} & Translation & Single & 101 & - & -  \\
\textsc{Belebele}~\citep{bandarkar-etal-2024-belebele} & Reading & Single & \textbf{122} & - & -  \\
\textsc{Include}~\citep{romanouinclude} & Knowledge & Single & 44 & - & -  \\
\textsc{X-AlpacaEval}~\citep{zhang2024plug} & Instruction & Single & 4 & 4 & - \\
\textsc{OMGEval}~\citep{liu2024omgeval} & Instruction & Single & 9 & 9 & -  \\
\textsc{SAGE}~\citep{guo2025largelanguagemodelstruly} & Culture & Single & 2 & 2 & 2  \\
\textsc{CultureScope}~\citep{zhang2025culturescope} & Culture & Single & 2 & 2 & 2  \\
\midrule
 \textbf{GaoYao (Ours)} & \textbf{Comprehensive} & \textbf{All (9 Sub-layers)} & 26 & \textbf{19} & \textbf{34}  \\
\bottomrule
\multicolumn{7}{l}{\footnotesize \dag \textit{Subj. Langs} refers to languages supported for open-ended subjective tasks (\emph{e.g.}, Instruction Following \& Multi-turn Dialogue).} \\
\end{tabular}
}
\caption{Comparison between GaoYao and other representative benchmarks. GaoYao distinguishes itself by its comprehensive scope, specifically surpassing others in subjective task coverage (19 languages) and cultural breadth (34 cultures) with rigorous human verification.}
\label{tab:comparison}
\end{table*}

\subsection{Evaluation of Multilingual and Multicultural Capabilities}
Evaluating the capabilities of LLMs across diverse languages and cultures is critical for their global deployment. Traditional evaluations primarily focus on MT and objective understanding tasks. \citet{goyal2022flores} introduced \textsc{Flores-101}, a benchmark covering massive languages for MT, while the WMT shared tasks~\citep{kocmi2024findings, barrault2019findings} remain the standard for evaluating translation quality using metrics like COMET~\citep{rei-etal-2020-comet, zouhar2024pitfalls}. Beyond translation, recent benchmarks have expanded to broader understanding capabilities. \textsc{Belebele}~\citep{bandarkar-etal-2024-belebele} evaluates parallel reading comprehension across 122 language variants, and \textsc{Include}~\citep{romanouinclude} assesses multilingual understanding with regional knowledge. Similarly, \citet{pomerenke2025ai} track LLM progress on various multilingual benchmarks.

However, existing evaluations often exhibit limitations in scope and depth. Many benchmarks rely heavily on objective formats such as multiple-choice questions (MCQs), neglecting subjective generation tasks that better reflect real-world usage. While \textsc{AlpacaEval}~\citep{alpaca_eval} and \textsc{MT-Bench}~\citep{zheng2023judging} provide robust evaluations for instruction following and dialogue, they are predominantly English-centric. Multilingual extensions like \textsc{OMGEval}~\citep{liu2024omgeval} and \textsc{X-AlpacaEval}~\citep{zhang2024plug} exist but often suffer from limited language coverage. Furthermore, cultural evaluation remains under-explored. \citet{yao2023benchmarking} and \citet{zhang2025culturescope} have initiated efforts to probe cultural awareness, and \citet{myung2024blend} introduced \textsc{BLEnD} for everyday cultural knowledge. Yet, these datasets often suffer from limited culture coverages, and the problem of relying on direct translation which retains source-culture bias still exists~\citep{rystrom2025multilingual, guo2025largelanguagemodelstruly}.

\subsection{Comparison with Existing Benchmarks.} 
As summarized in Table~\ref{tab:comparison}, existing benchmarks typically exhibit a trade-off between breadth and depth. 
Massive-scale benchmarks (\emph{e.g.}, \textsc{Flores-101}~\cite{goyal2022flores}, \textsc{Belebele}~\cite{bandarkar-etal-2024-belebele}) cover over 100 languages but are restricted to objective tasks (Translation and Reading). 
Conversely, benchmarks focusing on subjective tasks (\emph{e.g.}, \textsc{X-AlpacaEval}~\cite{zhang2024plug}) are severely limited in language coverage ($<10$). 
Crucially, even dedicated cultural benchmarks like \textsc{SAGE}~\cite{guo2025largelanguagemodelstruly} and \textsc{CultureScope}~\cite{zhang2025culturescope} are constrained to specific bilingual dyads (2 languages, 2 cultures).
Compared with these existing works, GaoYao establishes a more comprehensive framework. It integrates objective tasks with rigorously localized subjective tasks across 19 languages. Moreover, it significantly expands cultural evaluation through \textsc{SuperBLEnD}, a human-verified dataset covering 34 cultures, addressing the critical gap in deep, authentic multicultural assessment.

\subsection{LLMs for Multilingual and Multicultural Tasks}
The multilingual abilities of early LLMs are relatively underdeveloped~\cite{lai2024llms} due to the predominance of English in their pretraining data, such as the early LLaMA series~\cite{touvron2023llama}, \emph{e.g.}, the ratio of non-English languages in pretraining corpus of LLaMA-2~\cite{touvron2023llama2} is merely around 2\%. To improve the multilingual abilities of existing foundation LLMs, researchers have explored specialized tuning strategies. Many methods fouces on improving the quantity and quality of post-training using curated multilingual dataset and refined training strategies~\cite{chen2024monolingual,zhu2023extrapolating,zhang2024plug}, while others leveraged continual pre-training~\citep{fujiicontinual, nakamura2025aurora} and sparse Mixture-of-Experts architectures~\citep{zhu2025overcoming} to mitigate catastrophic forgetting~\citep{li2024revisiting}.

More recently, the focus has shifted towards monolingual capabilities into multilingual tasks. The ratio of multilingual data in pre-training corpus of recent LLMs is significantly growing~\citep{yang2025qwen3,deepseek2025v31}. In specific tasks (\emph{e.g.}, automatic speech recognition), the supported number of languages reaches 1600+~\cite{omnilingual2025omnilingual}. Despite the dramatic expansion of language coverage, the community has raised questions on the native level of responses in subjective tasks~\cite{liu2025midb} and authentic cultural experience for global users~\cite{rystrom2025multilingual}, where existing LLMs still lag behind. The three curated subjective and cultural test sets in GaoYao can serve as a pioneering step for bridging the gap for existing multilingual LLM to achieve truly equitable AI access.

\section{Detailed Information of Languages Supported by GaoYao}\label{sec:language_detail_info}

\begin{figure}[tbp]
  \centering
  \includegraphics[width=0.9\linewidth]{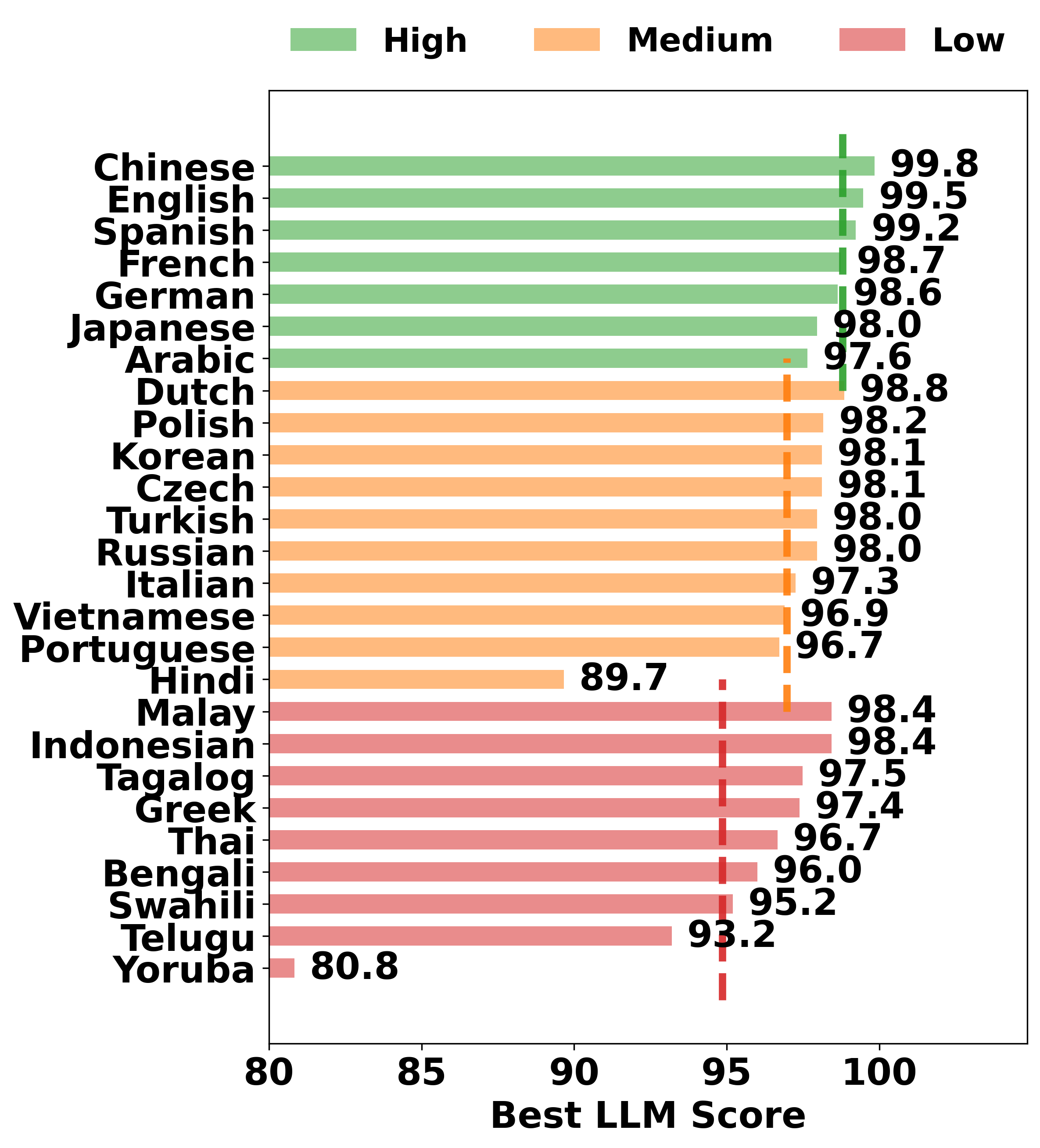}
\caption{Impact of language resource popularity on best performance achieved by LLMs. Vertical dashed lines represent group averages. The taxonomy on language resource-levels is based on~\citet{joshi2020state}. }
\label{fig:langage_distribution_scores_popularity}
\end{figure}

Table~\ref{tab:languages_table} presents the languages in our dataset, mapping full names to short codes, and detailing their geographical and resource-level information. Resource levels are determined using the \citet{joshi2020state} taxonomy, which rates languages from a set of 2,485 on a scale reflecting resource availability. According to this scale, we classify languages scoring 5 as high-resource (\emph{e.g., Chinese}), a score of 4 as mid-resource (\emph{e.g.}, Turkish), and a score of $\leq$3 as low-resource. Fig.~\ref{fig:langage_distribution_scores_popularity} displays the impact on LLM's best scores caused by resource popularity of languages, suggesting that LLM capabilities for languages with lower resources are relatively underdeveloped compared with high-resource ones.

\section{Reliability Analysis of GaoYao} \label{sec:reliability}

A robust benchmark must strike a balance: it should reflect current user needs (ecological validity) while probing capabilities that users may not yet explicitly request but are essential for advanced intelligence (comprehensive coverage). To assess this, we compared the topic distribution of GaoYao against 140k authentic user queries sampled from \textit{LMSYS Chatbot Arena}~\citep{chiang2024chatbot}. We employed a multilingual tagging model to categorize both datasets and calculated the Tag Semantic Alignment (TSA) score, \emph{i.e.}, the average semantic similarity between tags extracted from two groups.

Table~\ref{tab:dataset_rankings_simplified_v2} presents the results. We observe a \textbf{High Alignment in Utility Tasks}: The \textit{Instruction Following} (TSA 0.89) and \textit{Dialogue} (TSA 0.79) sub-layers show strong correlation with real-world queries, where tags like "IT Technology" and "Creative Writing" dominate. This confirms that our localization of subjective benchmarks accurately captures the core interaction patterns of global users.

Conversely, we see \textbf{Low Alignment in Specialized Domains}: Layers like \textit{Math} (0.03) and \textit{Cross-Cultural SA} (0.13) show lower alignment. This is likely because real-world users currently employ LLMs less frequently for complex tasks like advanced logic proofs or nuanced cultural philosophy. However, these "long-tail" capabilities are precisely where SOTA models differentiate themselves. By including these low-TSA but high-value domains, GaoYao prevents overfitting to "average" user behavior and ensures models are also evaluated on the frontier of capability.

\begin{table*}[tbp]
\centering
\resizebox{0.75\linewidth}{!}{
\begin{tabular}{lcll}
\toprule
Language & Code & Geographical Group & Resource Level \\
\midrule
English & EN & Western Europe & High Resource \\
French & FR & Western Europe & High Resource \\
German & DE & Western Europe & High Resource \\
Spanish & ES & Western Europe & High Resource \\
Dutch & NL & Western Europe & Medium Resource \\
Italian & IT & Western Europe & Medium Resource \\
Portuguese & PT & Western Europe & Medium Resource \\
Greek & EL & Western Europe & Low Resource \\
Czech & CS & Eastern Europe & Medium Resource \\
Polish & PL & Eastern Europe & Medium Resource \\
Russian & RU & Eastern Europe & Medium Resource \\
Chinese & ZH & East Asia \& Southeast Asia & High Resource \\
Japanese & JA & East Asia \& Southeast Asia & High Resource \\
Korean & KO & East Asia \& Southeast Asia & Medium Resource \\
Vietnamese & VI & East Asia \& Southeast Asia & Medium Resource \\
Indonesian & ID & East Asia \& Southeast Asia & Low Resource \\
Malay & MS & East Asia \& Southeast Asia & Low Resource \\
Tagalog & TL & East Asia \& Southeast Asia & Low Resource \\
Thai & TH & East Asia \& Southeast Asia & Low Resource \\
Arabic & AR & Middle East \& Africa & High Resource \\
Turkish & TR & Middle East \& Africa & Medium Resource \\
Swahili & SW & Middle East \& Africa & Low Resource \\
Yoruba & YO & Middle East \& Africa & Low Resource \\
Hindi & HI & South Asia & Medium Resource \\
Bengali & BN & South Asia & Low Resource \\
Telugu & TE & South Asia & Low Resource \\
\bottomrule
\end{tabular}
}
\caption{Language mapping: codes, geographical groups, and resource levels.}
\label{tab:languages_table}
\end{table*}

\begin{table*}[htbp]
\centering
\renewcommand{\arraystretch}{1.3}
\small
\begin{tabularx}{\textwidth}{@{}c >{\RaggedRight\arraybackslash}p{1.8cm} c >{\RaggedRight\arraybackslash}X@{}}
\toprule
\textbf{ } & \textbf{Sub-layer} & \textbf{TSA} & \textbf{Top-10 Frequent Tags} \\
\midrule
1 & Cross-Cultural (SA) & 0.1323 & Educational Philosophy, Philosophy/Religion, Values, Higher Education, Language/Writing, Humanities, Workplace Life, Politics \\
\midrule
2 & Cross-Cultural (SB) & 0.4548 & Workplace Life, Social Customs, Food \& Cooking, IT Technology, Higher Education, Sports, Artifacts, AI \\
\midrule
3 & Mono-Cultural (CS) & 0.2116 & Workplace Life, Social Customs, Educational Philosophy, Humanities, Language/Writing, Business Management, Values, Banking \\
\midrule
4 & Math & 0.0319 & Applied Math, Logic, Algebra, Business Management, Probability, Operations Research, Education Info \\
\midrule
5 & Reasoning & 0.3382 & Philosophy/Religion, Civil Law, Politics, World History, Business Management, Economics, Constitution, Clinical Medicine \\
\midrule
6 & Knowledge & 0.3297 & World History, Geography, Business Management, Physiology, Ecology, Economics, Zoology, Politics \\
\midrule
7 & Instruct Follow & 0.8955 & IT Technology, Food \& Cooking, Language/Writing, Literature, Movies/TV, AI, Music, Games, Business Management \\
\midrule
8 & Dialogue & 0.7980 & IT Technology, Literature, Algebra, Language/Writing, Business Management, Probability, Movies/TV, AI, Physics \\
\midrule
9 & Reading & 0.5046 & World History, Geography, IT Technology, Politics, Travel, Zoology, Transportation, Social Customs \\
\midrule
10 & Translation & 0.2117 & Artifacts, Values, Language/Writing, Regional Characteristics, Zoology, Geography, World History, Politics \\
\bottomrule
\end{tabularx}
\caption{Tag Semantic Alignment (TSA) scores comparing GaoYao layers with real-world user queries (LMArena). High TSA indicates alignment with common daily usage; low TSA indicates specialized or long-tail capabilities.}
\label{tab:dataset_rankings_simplified_v2}
\end{table*}

\section{Data Card for Human Annotation}
\label{sec:data_card}

Annotators were selected based on native proficiency and professional experience in translation, localization, proofreading, editing, copy-writing, technical writing, or linguistic testing. The annotation guidelines emphasized meaning-preserving localization rather than literal translation, especially for instructions involving scripts, phonetics, culturally bound terms, or pragmatic conventions. For \textsc{SuperBLEnD}, annotators were instructed to answer from lived cultural experience without using AI systems or search engines, and to mark a question as "not applicable" when no clear cultural consensus existed. All annotations followed a review-rebuttal protocol: third-party reviewers flagged questionable cases, after which annotators either revised the answer or provided a justification for keeping it. Annotators were full-time employees of the partner language service center and participated as part of their regular professional duties, receiving standard salaries above local minimum wage rather than unpaid or crowdsourced compensation.

\section{Technical Details in the Curation of \textsc{SuperBLEnD}}\label{sec: detailed superblend annotation}

As discussed in Section~\ref{sec:integration}, existing cultural evaluation sets are limited in their culture coverage. However, unlike the two multilingual abilities in Section~\ref{sec:expansion}, expanding coverage of existing cultural evaluation sets presents a unique challenge: direct translation of cultural QA pairs retains the cultural perspective of the source language (especially for the answer parts) which fails to test true cross-cultural generalizability, while a purely manual reconstruction is prohibitive in costs. 

To address this, we developed a three-stage semi-automated data expansion procedure incorporating human verification, where native speakers provide initial seeds in the first stage and inspect quality in subsequent diversifying stages to ensure both high accuracy and linguistic diversity. The resulting evaluation set, denoted as \textsc{SuperBLEnD}, is a generalization of the \textsc{BLEnD} dataset and evaluates LLMs' understanding of cultural differences regarding everyday concepts across 34 nations (up from 16 in \textsc{BLEnD}).

\paragraph{Stage 1: Generalization Q\&A Seeds to More Cultures} Starting from the question templates released by \textsc{BLEnD}, which cover culturally shared topics ranging from festivals and food to sports, we first selected a high-quality subset. This filtration process excluded templates which were not universally applicable or were deemed sensitive in specific cultural contexts. To ensure cultural authenticity and factual accuracy, answers to the selected questions were built by native members of corresponding cultures. These annotators come from the cooperated language service center as described in Section~\ref{sec:expansion}. Of the 34 cultures SuperBLEnD covers, data for 16 was inherited from \textsc{BLEnD}. For each of the 18 newly expanded cultures, three native annotators are assigned. The instruction asks annotators to provide 1-3 concise answers to each question based strictly on their personal life experiences within that cultural context, unassisted by AI or search engines. To prevent forced fabrication, annotators could mark questions as “not applicable” or “no clear answer.” 

Q\&A pairs across all 34 cultures underwent rigorous manual verification, discarding approximately 41.1\% of the raw data. The removed cases mainly fall into four categories. \textit{Semantic redundancy} covers near-duplicate answers that express the same concept, such as merging "Mum" and "Mother" into a single normalized answer. \textit{Context errors} refer to answers that are linguistically plausible but culturally or semantically incorrect in the target context; for example, "Wochenbett" was rejected for a German question about postpartum recovery locations because it refers to the puerperium period rather than a physical place. \textit{Hierarchical conflicts} occur when a distractor or answer overlaps with another valid answer at a different granularity; for instance, "Pepsi" is unsuitable when "carbonated drinks" is also a valid answer. \textit{Safety issues} include sensitive or potentially toxic responses that are inappropriate for a public benchmark. The final collection contains an average of 2.17 high-quality answers per question template. Note that many cultural questions accept multiple correct answers. For example, both "beer" and "carbonated drinks" are valid, verified responses to the question: "What do young people in Malaysia usually drink at nightclubs?"

\paragraph{Stage 2: Option Synthesis} For ease of evaluation and to enhance diversity, following \citet{myung2024blend}, we generalize each verified seed question into a series of multiple-choice questions (MCQs). The volume of generated MCQs is dynamically adjusted based on the answer set size, ensuring that all verified answers are comprehensively covered as correct options while upsampling (with different distractors) questions with fewer answers as a balance. For each MCQ, we synthesized options by combining the correct answer for the target country with three wrong answers as distractors derived from other countries. In cases where insufficient real-world distractors were available, an LLM was employed to generate plausible but incorrect "dummy options" that exist in reality but do not answer the specific question with the following prompt:
\begin{mybox}
Provide \{3 - n\} dummy option(s) that makes sense to be the answer(s) of the given question, and has to exist in real-life (non-fiction), but is totally different from the given answers without any explanation. Make sure that the options are different from each other, and cannot be an answer from any country. Provide as JSON format: \{"dummy\_options":[]\} 
\end{mybox}

Synthesized MCQs underwent automated and human verification to ensure safety, answer uniqueness, and the exclusion of synonyms or hierarchical conflicts. For instance, in the “Malaysia nightclub” scenario, “Pepsi” is an unsuitable distractor when the target answer is “beer.” Because “Pepsi” falls under the category of “carbonated drinks” (another acceptable cultural answer), its inclusion introduces a hierarchical relationship that compromises the MCQ's validity.

\paragraph{Stage 3: Linguistic Enrichment} 

To further enhance linguistic diversity and reasoning difficulty, the finalized MCQs underwent a rephrasing stage. We utilized an LLM prompted to rewrite both the question stem and options, employing techniques like paraphrasing, voice alternation, and syntactic restructuring without altering the core semantic meaning or named entities. The prompt is as follows: 
\begin{mybox}
I will give you a multiple-choice cultural question. Your task: refine the wording of both the stem and the option as I required. Goal: raise the overall difficulty and enrich the phrasing while keeping the underlying concepts intact. Recommended techniques: paraphrasing, expansion, morphological variation, idioms and figurative language, voice alternation (active $\leftrightarrow$ passive), syntactic restructuring, etc. Requirements: 1. You must not change the semantic meaning of the original stem and options. 2. Do not alter any entities or proper nouns (e.g., personal names, company names, sport names, countries/region names, festival names). 3. Do not add a country or regional name (or its adjective) to the options unless the answer itself is a country or region. 4. Ensure the index of the original correct answer stays the same. 5. Use English. 6. Keep capitalization consistent across the options; capitalizing the first letter of each option is recommended. 7. Follow the specified output format exactly, including JSON punctuation. Output format: [...] 
\end{mybox}
This transforms straightforward MCQs into more complex versions while retaining the same cultural core, ensuring the benchmark tests cultural knowledge rather than simple pattern matching 

\section{Disclosure of Generative AI Usage}

The technology of generative AI is partially involved in this paper for the following three scenarios: (1) polishing texts for language fluency, (2) aiding the curation of SuperBLEnD dataset as specified in Appendix~\ref{sec: detailed superblend annotation} and (3) aiding in plotting art elements in figures and diagrams (\emph{e.g.}, the iceberg icon in Fig.~\ref{fig:main_methodology}).

\section{Detailed Experimental Setups}\label{sec:detailed evaluation approaches}

\begin{table*}[tbp]
    \centering
    
    \resizebox{0.9\linewidth}{!}{
    \setlength{\tabcolsep}{2pt} 
    \begin{tabularx}{\textwidth}{@{}l p{1.2cm} c c p{1cm} p{1.3cm} X@{}}
        \toprule
        \textbf{Data source} & \textbf{Task Type} & \textbf{Eval. Type} & \textbf{Metric} & \textbf{Judge Model} & \textbf{Ref. Source} & \textbf{Calculation Method} \\
        \midrule
        S-AlpacaEval & QA & Subj. & Win Rate & Deep Seek V3.1 & Qwen3-235B & 1. Judge compares candidate vs. reference (correctness, richness, comprehensiveness, etc.); \newline 2. $\text{Win Rate} = \frac{\#\text{win} + \#\text{tie}/2}{\#\text{all}}$ \\
        \midrule
        Belebele & MCQ & Obj. & Accuracy & Rule-based & Human (Open Source) & 1. Reading comprehension, 4-option regex match (A-D); \newline 2. $\text{Accuracy} = \frac{\#\text{correct}}{\#\text{all}}$ \\
        \midrule
        INCLUDE & MCQ & Obj. & Accuracy & Rule-based & Human (Open Source) & 1. Encyclopedic knowledge, 4-option regex match (A-D); \newline 2. $\text{Accuracy} = \frac{\#\text{correct}}{\#\text{all}}$ \\
        \midrule
        SuperBLEnD & MCQ & Obj. & Accuracy & Rule-based & Human and LLM Hybrid  & 1. Regional culture knowledge, 4-option regex match (A-D); \newline 2. $\text{Accuracy} = \frac{\#\text{correct}}{\#\text{all}}$ \\
        \midrule
        MGSM & Math & Obj. & Accuracy & Rule-based & Human (Open Source) & 1. Math reasoning, regex match for integer answers; \newline 2. $\text{Accuracy} = \frac{\#\text{correct}}{\#\text{all}}$ \\
        \midrule
        MMMLU & MCQ & Obj. & Accuracy & Rule-based & Human and LLM Hybrid (Open Source) & 1. Knowledge QA, 4-option regex match (A-D). Uses LLM if regex fails; \newline 2. $\text{Accuracy} = \frac{\#\text{correct}}{\#\text{all}}$ \\
        \midrule
        Flores-101 & Translation & Obj. & Comet & wmt22 comet -da  & Human Translated Wiki & 1. Translation task; \newline 2. Comet Score \\
        \midrule
        SAGE & MCQ+ T/F+ QA & \makecell[l]{Subj.+ Obj.} & Mixed & Deep Seek V3.1 & Qwen3-max & 1. MCQ and T/F uses accuracy as score; 2. QA use LLM to recognize culture points mentioned in the answer; 3. weighted sum score is used as final score.\\
        \midrule
        CultureScope & MCQ+ T/F+ QA & \makecell[l]{Subj.+ Obj.} & Mixed & Deep Seek V3.1 & Human Expert & 1. MCQ and T/F uses accuracy as score; 2. QA use LLM to recognize culture points mentioned in the answer; 3. weighted sum score is used as final score. \\
        \midrule
        S-MT-Bench & QA & Subj. & Win Rate & Deep Seek V3.1 & Qwen3-235B-A22B & 1. Judge comparison (multi-turn averaged); \newline 2. $\text{Win Rate} = \frac{\#\text{win} + \#\text{tie}/2}{\#\text{all}}$ \\
        \bottomrule
    \end{tabularx}

    }
    \caption{Summary of evaluation methodologies. Task types include Multiple Choice Questions (MCQ), True/False (T/F), and Open-ended Q\&A (QA). Evaluation types distinguish between subjective (Subj.) LLM-judged approaches and objective (Obj.) rule-based approaches. MGSM only generate Integer as final answer.}
    \label{tab:evaluation_method}
    
\end{table*}
\begin{table*}[tbp]
\centering
\scriptsize 
\renewcommand{\arraystretch}{1.2} 
\begin{tabularx}{\textwidth}{@{}l l X@{}}
\toprule
\textbf{Model} & \textbf{Model Version} & \textbf{Resource Address} \\
\midrule
\multicolumn{3}{c}{\textit{Flagship Open-Source Models}} \\
\midrule
openPangu-Ultra-MoE-718B-V1.1\cite{openpangu2024ultra} & openPangu-Ultra-MoE-718B-V1.1 & \url{https://ai.gitcode.com/ascend-tribe/openPangu-Ultra-MoE-718B-V1.1} \\
DeepSeek-V3.1\cite{deepseek2025v31} & DeepSeek-v3.1-250821 & \url{https://huggingface.co/deepseek-ai/DeepSeek-V3.1} \\
Qwen3-235B-A22B\cite{yang2025qwen3} & Qwen3-235B-A22B-Instruct-2507 & \url{https://huggingface.co/Qwen/Qwen3-235B-A22B-Instruct-2507} \\
Qwen3-VL-235B-A22B\cite{bai2025qwen3vltechnicalreport} & Qwen3-VL-235B-A22B-Instruct & \url{https://huggingface.co/Qwen/Qwen3-VL-235B-A22B-Instruct} \\
DeepSeek-R1\cite{deepseek2025r1} & DeepSeek-R1 & \url{https://huggingface.co/deepseek-ai/DeepSeek-R1} \\
Llama-3.1-405B\cite{meta2024llama31} & Llama-3.1-405B-Instruct & \url{https://huggingface.co/meta-llama/Llama-3.1-405B-Instruct} \\
GLM-4.6\cite{5team2025glm45agenticreasoningcoding} & GLM-4.6 & \url{https://chatglm.cn} \\
Kimi-k2\cite{kimiteam2025kimik2openagentic} & kimi-k2-250711 & \url{https://www.kimi.com} \\
\midrule
\multicolumn{3}{c}{\textit{Closed-Source Commercial Models}} \\
\midrule
Doubao-seed-1.6\cite{seed1_6} & doubao-seed-1-6-250615 & \url{https://www.doubao.com/chat} \\
Qwen-max\cite{qwen3max} & Qwen-max & \url{https://chat.qwen.ai/} \\
Gemini-2.5-Pro\cite{comanici2025gemini} & Gemini-2.5-Pro & \url{https://aistudio.google.com} \\
Claude-Sonnet-4.5\cite{anthropic2026claude} & claude-sonnet-4-5-20250929 & \url{https://claude.ai} \\
Grok-3\cite{xai2025grok3} & Grok-3 & \url{https://grok.com} \\
o3\cite{openai2025o3o4mini} & o3 & \url{https://chatgpt.com} \\
o4-mini\cite{openai2025o3o4mini} & o4-mini & \url{https://chatgpt.com} \\
GPT-5-chat\cite{openai2026gpt5} & GPT-5-chat & \url{https://chatgpt.com/} \\
GPT-4o\cite{openai2024gpt4ocard} & GPT-4o & \url{https://chatgpt.com} \\
\midrule
\multicolumn{3}{c}{\textit{Compact Models (<20B)}} \\
\midrule
Qwen3-14B\cite{yang2025qwen3} & Qwen3-14B & \url{https://huggingface.co/Qwen/Qwen3-14B} \\
Qwen3-8B\cite{yang2025qwen3} & Qwen3-8B & \url{https://huggingface.co/Qwen/Qwen3-8B} \\
Gemma-3-12B-IT\cite{gemmateam2025gemma3technicalreport} & gemma-3-12b-it & \url{https://huggingface.co/google/gemma-3-12b-it} \\
Llama-3.1-8B\cite{meta2024llama31} & Llama-3.1-8B-Instruct & \url{https://huggingface.co/meta-llama/Llama-3.1-8B-Instruct} \\
Llama-3-8B\cite{meta2024llama31} & Llama-3-8B-Instruct & \url{https://huggingface.co/meta-llama/Meta-Llama-3-8B-Instruct} \\
Ministral-8B-Instruct\cite{jiang2024mixtralexperts} & Ministral-8B-Instruct-2410 & \url{https://huggingface.co/mistralai/Ministral-8B-Instruct-2410} \\
\bottomrule
\end{tabularx}
\caption{Detailed specifications of models evaluated in GaoYao.}
\label{tab: model addresses}
\end{table*}

\end{document}